%% file: main.tex
\documentclass[letterpaper, 10 pt, conference]{ieeeconf}  

\IEEEoverridecommandlockouts                              

\usepackage{graphics} 
\usepackage{epsfig} 
\usepackage{mathptmx} 
\usepackage{times} 
\usepackage{amsmath} 
\usepackage{amssymb}  
\usepackage[caption=false,font=normalsize,labelfont=sf,textfont=sf]{subfig}
\usepackage{textcomp}
\usepackage{stfloats}
\usepackage{url}
\usepackage{verbatim}
\usepackage{graphicx}
\usepackage{cite}
\usepackage[utf8]{inputenc}
\usepackage{caption}
\usepackage{mathrsfs}
\usepackage{amssymb}
\usepackage{multirow}
\usepackage{csquotes}
\usepackage{booktabs}
\usepackage{makecell}
\usepackage{float}
\floatstyle{plaintop}
\restylefloat{table}
\usepackage[colorlinks,linkcolor=red,anchorcolor=blue,citecolor=green]{hyperref}

\title{\LARGE \bf
CrackNex: a Few-shot Low-light Crack Segmentation Model Based on Retinex Theory for UAV Inspections}

\author{Zhen Yao$^{1}$, Jiawei Xu$^{1}$, Shuhang Hou$^{1}$ and Mooi Choo Chuah$^{1}$
\thanks{*This work was supported by National Science Foundation Grant CPS 1931867.}
\thanks{$^{1}$All authors are with the Department of Computer Science and Engineering, P.C. Rossin College of Engineering and Applied Science, Lehigh University, Bethlehem, PA 18015, USA. \{zhy321, jix519, shh420, mcc7\}@lehigh.edu}
}

\begin{document}

\maketitle
\thispagestyle{empty}
\pagestyle{empty}

\input{Body/1_Intro}
\input{Body/2_RelatedWork}
\input{Body/3_Method}
\input{Body/4_Results}

\bibliographystyle{IEEEtran}
\bibliography{refer}
\end{document}

%% file: Body/1_Intro.tex
\begin{abstract}
Routine visual inspections of concrete structures are imperative for upholding the safety and integrity of critical infrastructure. Such visual inspections sometimes happen under low-light conditions, e.g., checking for bridge health. Crack segmentation under such conditions is challenging due to the poor contrast between cracks and their surroundings. However, most deep learning methods are designed for well-illuminated crack images and hence their performance drops dramatically in low-light scenes. In addition, conventional approaches require many annotated low-light crack images which is time-consuming. In this paper, we address these challenges by proposing CrackNex, a framework that utilizes reflectance information based on Retinex Theory to learn a unified illumination-invariant representation. Furthermore,  we utilize few-shot segmentation to solve the inefficient training data problem. In CrackNex, both a support prototype and a reflectance prototype are extracted from the support set. Then, a prototype fusion module is designed to integrate the features from both prototypes. CrackNex outperforms the SOTA methods on multiple datasets. Additionally, we present the first benchmark dataset, LCSD, for low-light crack segmentation. LCSD consists of 102 well-illuminated crack images and 41 low-light crack images. The dataset and code are available at \href{https://github.com/zy1296/CrackNex}{https://github.com/zy1296/CrackNex}. \par
\end{abstract}

\IEEEpeerreviewmaketitle

\section{Introduction\label{intro}}
Cracks are common defects on pavement and in concrete structures. Overloading, structural changes, and environmental hazards may accelerate these deteriorations, causing a significant safety risk \cite{yan2022cycleadc}. Therefore, regular inspection of roads and bridges to identify damage and repair defects is essential to maintain building and traffic safety. In recent years, a variety of deep-learning algorithms \cite{choi2019sddnet,liu2019deepcrack,kang2020hybrid,liu2020automated,rezaie2020comparison,zheng2023robustness, hsieh2021neural, wang2024landa, zhu2023topic} have been proposed. \par

However, these algorithms exhibited limited effectiveness in real-world scenarios due to variations in lighting conditions, the presence of shadows, and other factors \cite{wang2022automatic,fan2023pavement,li2019crack}. In real-life scenarios, there exist numerous instances of low-light conditions where distinguishing cracks becomes challenging. For example, cracks on the underside of bridge piers, in tunnel walls, and in historical buildings are in remote or hard-to-reach areas that have limited natural light. Engineers often rely on artificial lighting, which may not provide optimal visibility and may cause information loss. Thus, computer vision-based low-light crack segmentation is necessary for safety inspections. \par

\begin{figure}[!t]
\centering
\includegraphics[width=\linewidth]{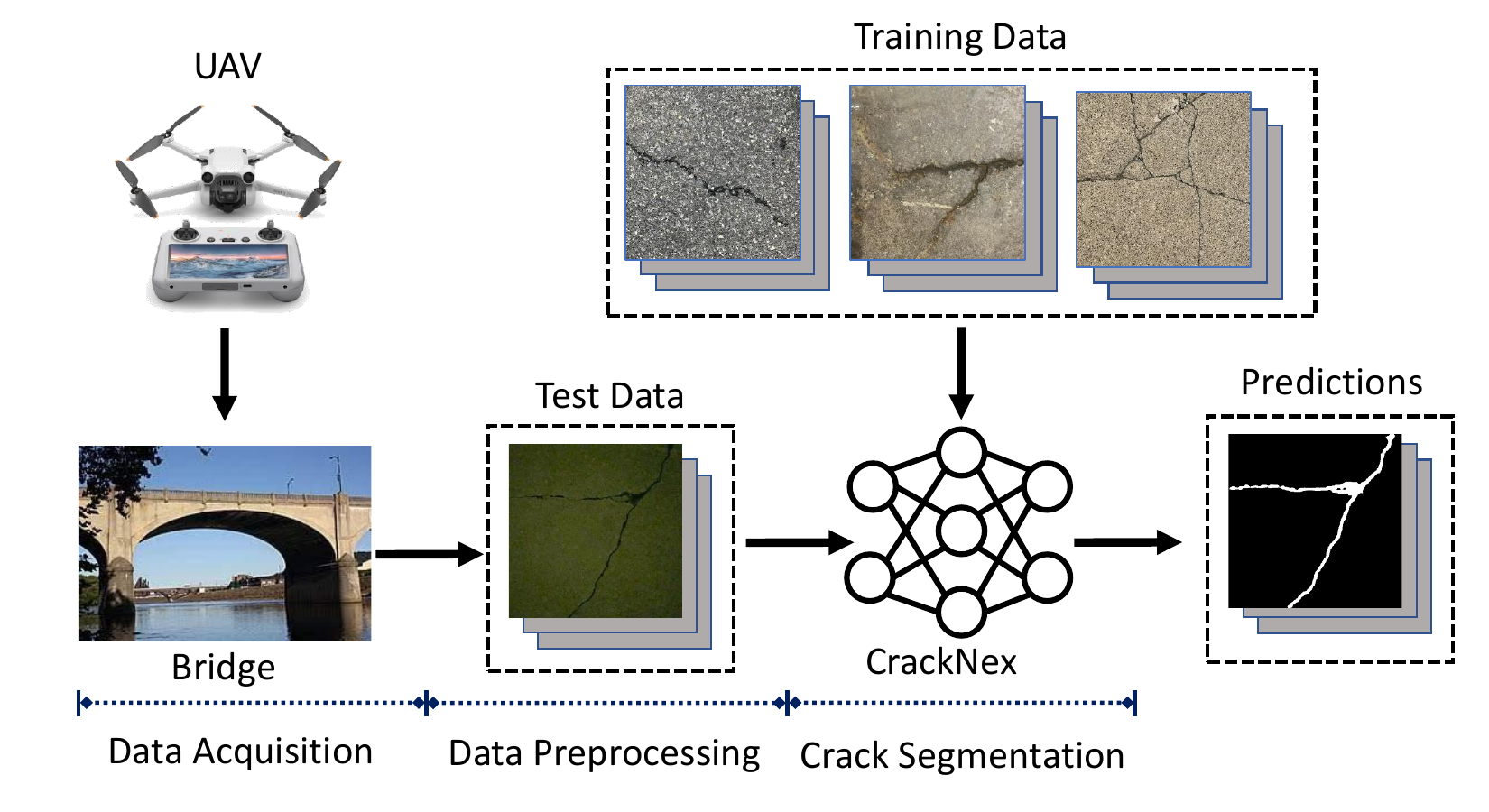}
\caption{System overview for UAV inspections}
\label{system}
\end{figure}

Low-light crack segmentation is challenging because most deep-learning methods were trained on normal-light crack images captured under well-lit conditions \cite{li2021automatic,wang2021renet,dong2022innovative}. In unfavorable low-light conditions, the segmentation performance drops significantly, suffering from both low image contrast and ambiguity of object boundaries. Therefore, we introduce the Retinex Theory \cite{land1977retinex} to address this challenge. It, rooted in human color perception, posits that the observed color image can be decomposed into reflectance and illumination. Reflectance represents the intrinsic attributes of captured objects, which remain consistent under different light conditions. Illumination describes the luminance values present on objects. By using a pre-trained Decompose Network \cite{wei2018deep}, we can estimate the reflectance features and help the model learn a unified and illumination-invariant representation. \par

Additionally, deep learning is essentially data-driven while collecting such low-light crack images and making high-quality annotations at pixel level are not only time-consuming but also prone to human errors. To tackle this challenge (scarcity of annotated low-light data), we utilize the few-shot segmentation method. The increasingly popular few-shot learning is a promising direction to address the limitation of insufficient data by training models to generalize to new classes with a small number of examples (or shots) \cite{zhao2021few}. This is particularly valuable in situations where acquiring abundant data is difficult. Unlike common deep-learning approaches, few-shot segmentation has better generalization ability which makes it more adaptable to new or unseen classes \cite{wang2019panet,dong2018few,min2021hypercorrelation,mao2022learning, sun2022few, pan2023learning}. In few-shot crack segmentation, our goal is to train a model using a sufficient number of normal-light crack images and during inferring, this trained model can be used to segment low-light cracks given a few labeled low-light crack images, as illustrated in Figure 1. \par

In this work, we propose a reflectance-guided few-shot low-light crack segmentation network, CrackNex. It leverages reflectance to enhance the learning of contrast and recover lost details in low-light images. In addition to extracting the support prototype from the support set, we generate a complementary reflectance prototype from the reflectance features. The prototype fusion module (PFM) is proposed to integrate the support prototype with the reflectance prototype and it is capable of uncovering the inner connection between two prototypes with attention weights learned through the co-attention mechanism. Moreover, an Atrous Spatial Pyramid Pooling (ASPP) module \cite{chen2017rethinking} is applied to improve the features extracted by the CNN backbone in a multi-scale manner. \par

In summary, our contributions to this paper include:
\begin{itemize}
\item To the best of our knowledge, we propose the first few-shot method for low-light crack segmentation.
\item We introduce reflectance information from Retinex Theory and propose a novel reflectance-guided network, CrackNex. Our work highlights a new direction for low-light segmentation. Reflectance indicates the intrinsic properties of different objects and can help distinguish cracks from other non-crack regions. 
\item We conduct experiments to evaluate our CrackNex model on two crack segmentation datasets and demonstrate its effectiveness using standard segmentation metrics. Compared to several SOTA models, our CrackNex achieves the SOTA performance on these 2 datasets.
\item We present a new crack segmentation dataset, LCSD, with both well-illuminated and low-light crack images for the benefit of the research community.
\end{itemize}


%% file: Body/2_RelatedWork.tex
\section{Related works\label{related}}
\subsection{Few-Shot Learning}
Few-shot learning, the task of training models to recognize and generalize from a limited number of examples, has garnered significant attention in recent years due to its applicability in various domains. Existing methods are mainly meta-learning and metric learning. \par

Several meta-learning approaches \cite{finn2017model,garcia2017few,munkhdalai2017meta,wang2022metateacher,li2022domain} have been proposed to learn transferable knowledge from diverse learning tasks, leading to substantial advancements in the field. Elsken \cite{elsken2020meta} combined meta-learning with gradient-based Neural Architecture Search (NAS) \cite{zoph2016neural} methods and designed a meta-learning framework capable of customizing the meta-architecture to task-specific architectures. Baik \cite{baik2021meta} proposed a novel framework to learn a task-adaptive loss function through two meta-learners, employing two distinct meta-learners: one responsible for learning the loss function and another for learning parameters for the loss function. \par

Metric learning \cite{ying2021weakly} leverages distance measurement to optimize the distance or similarity between the images and regions. Fan \cite{fan2022self} proposed a novel self-support network by leveraging self-support matching to solve the appearance discrepancy problem. Okazawa \cite{okazawa2022interclass} proposed a novel few-shot segmentation approach that effectively enhanced the distinction between the target class and closely resembling classes, yielding improved separation performance. Our work is inspired by the metric-based approach to generate better prototypes utilizing reflectance information. \par

\subsection{Crack Segmentation}
Semantic segmentation, a pixel-level image classification task, is essential for understanding and interpreting visual data. The deep-learning method \cite{choi2019sddnet,liu2019deepcrack,lau2020automated,konig2021optimized,qiu2023sats,wang2022automatic,fan2023pavement} has shown promising results in the crack segmentation task. \par

Early works rely on conventional semantic segmentation models. Liu \cite{liu2019deep} applied U-Net for pavement crack segmentation and proposed an unmanned aerial system for UAV inspections \cite{sun2023hmaac,wei2022lidar,dang2024enhancing,ma2023implementation}. Liu \cite{liu2020automated} extended the U-Net architecture by incorporating additional convolutional layers to develop a pavement crack segmentation method. Sarmiento \cite{sarmiento2021pavement} used another successful segmentation model, DeepLabv3, to segment pavement cracks. \par

Recently, Transformers and attention mechanisms have also been widely used for crack segmentation. Wang \cite{wang2022automatic} designed a segmentation model that utilizes a hierarchical Transformer as the encoder and integrates a top-down structure. Xiang \cite{xiang2023crack} introduced a dual encoder–decoder model by using both transformers and CNNs to achieve precise segmentation of crack images. \par

%% file: Body/3_Method.tex
\begin{figure*}[!t]
\centering
\includegraphics[width=1.00\textwidth]{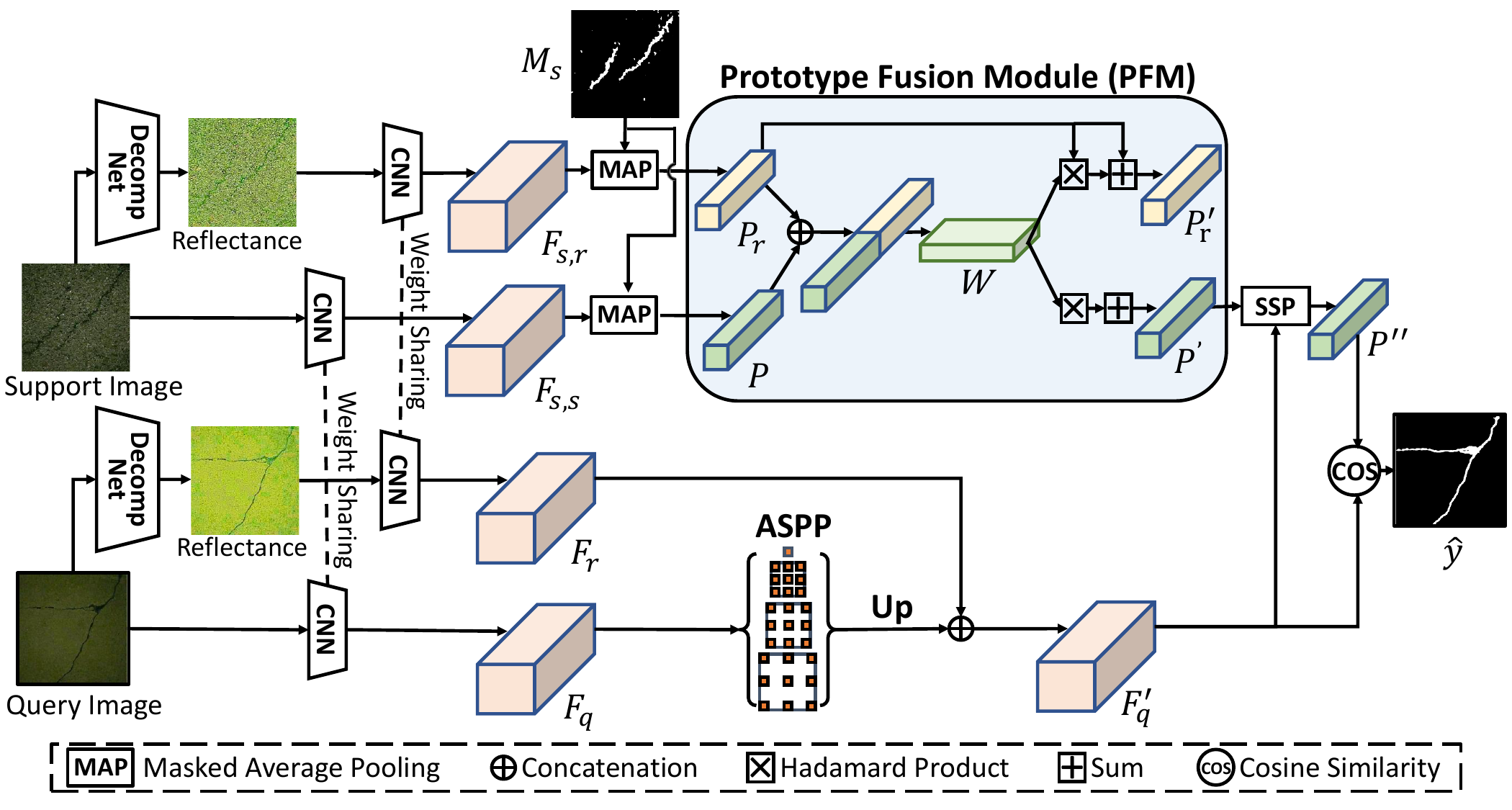}
\caption{Illustration of CrackNex: We generate reflectance features on query image and support images respectively by using Decompose Net. Afterwards, we generate the support and reflectance prototypes and update both prototypes by using the Prototype Fusion Module. Meanwhile, reflectance query features are utilized as low-level features in the Atrous Spatial Pyramid Pooling (ASPP) module to preserve details. Finally, we further update the support prototype in the Self-Support Prototype (SSP) module to perform matching with query features and calculate the loss.}
\label{model}
\end{figure*}

\section{Our proposed low-light crack segmentation model\label{method}}
\subsection{Problem Formulation}
Assessing structural integrity and identifying potential damages of infrastructure such as bridges or old buildings is a critically important task for repairs that can be carried out in-time to ensure the safety of human lives. However, sometimes, limited budgets or lack of workers may prevent such task from being carried out as frequently as one wishes. With the recent emergence of affordable drones, such inspections can be conducted effectively without much cost. In Fig.\ref{system}, we show an example of how Unmanned Aerial Vehicles (UAV) can be used to inspect the lateral side and underside of bridges. \par

\subsection{Background}
Few-shot segmentation aims to generate pixel-level segmentation predictions of novel classes when only a limited number of annotated labels are available. The setup is defined in an $N$-ways-$K$-shot format, where $N$ represents the number of classes, and $K$ indicates the number of support images needed for a query image. Our crack segmentation task involves learning to predict 1-way-1-shot and 1-way-5-shot segmentations. In the 1-shot setting, the model uses one single support image as a reference, and in the 5-shot setting, it utilizes 5 support images to generate predictions. The foreground refers to cracks, while the background is non-crack regions. \par

We use SSP \cite{fan2022self} as our baseline few-shot segmentation model. Fan et al. proposed this self-support matching framework, which utilized query features to generate self-support prototypes. It addresses the intra-class appearance discrepancy problem in few-shot segmentation by effectively reducing the gap between support prototypes and query features. Such capability is useful for our work. \par

The training set contains well-illuminated crack images and we represent the normal-light crack as base class $\mathbf{C_{train}}$. Then, we evaluate the trained model using low-light crack images (novel class $\mathbf{C_{test}}$). Following a previous work \cite{ying2020}, an episode-based sampling strategy is employed during both training and evaluating. Specifically, every sampled episode $e_i = \{\mathbf{S_{i}}, \mathbf{Q_{i}}\}$ of each class $\mathbf{C}$ consists of the support set and the query set. The support set $\mathbf{S_{i}} = \{(\mathbf{I_{s}^{k}}, \mathbf{M_{s}^{k}}), k \in \{1,\dots,K\}\}_{i}$ consists of a collection of support images, where $\mathbf{I_{s}^{k}}$,$\mathbf{M_{s}^{k}}$ are the $k^{th}$ support image and its masks, and the query set $\mathbf{Q_{i}} = \{(\mathbf{I_{q}}, \mathbf{M_{q}})\}_{i}$ refers to a collection of images for which the model needs to perform segmentation, where $\mathbf{I_{q}}$ is the query image and its ground truth mask $\mathbf{M_{q}}$. \par

\subsection{Architecture Overview of CrackNex}
Our model consists of a Decompose Net, a Prototype Fusion module (PFM), an ASPP module \cite{chen2017rethinking}, and an SSP module as illustrated in Fig. \ref{model}. Our unique contribution includes utilizing reflectance information to improve segmentation performance and designing the PFM module to fuse extracted support features and reflectance features. The Prototype Fusion Module is described in Section \ref{pfm} and the ASPP module is described in Section \ref{aspp}. \par

We first use a Decompose Net from RetinexNet \cite{wei2018deep} pre-trained on the LOL dataset \cite{wei2018deep} to generate reflectance images on both support and query images. The Decompose Net is responsible for separating an input image into a reflectance image and an illumination image. The reflectance features can be further used to highlight object boundaries and capture details, e.g., color, texture, and surface characteristics, leading to more accurate edge and robust segmentation predictions. \par

We then apply two CNN backbones pre-trained with ImageNet \cite{deng2009imagenet} to extract feature maps $\mathbf{F_q} \in \mathbb{R}^{H/8 \times W/8 \times C}$ and $\mathbf{F_r} \in \mathbb{R}^{H/8 \times W/8 \times C}$ for the query image and query reflectance image, respectively. The image features are integrated with the reflectance features through the ASPP module at multiple scales. The output of the ASPP module is updated query features $\mathbf{F'_q} \in \mathbb{R}^{H/4 \times W/4 \times C}$. \par

Another two backbones are also applied to generate features on support images and support reflectance images. Note that the two backbones shared weights with the former backbones in pairs. Afterwards, extracted support features $\mathbf{F_{s,s}} \in \mathbb{R}^{H/8 \times W/8 \times C}$ and support reflectance features $\mathbf{F_{s,r}} \in \mathbb{R}^{H/8 \times W/8 \times C}$, along with ground truth mask $\mathbf{M_s}$, are fed into the masked average pooling layer to generate the support prototype $\mathbf{P} \in \mathbb{R}^{1 \times 1 \times C}$ and reflectance prototype $\mathbf{P_r} \in \mathbb{R}^{1 \times 1 \times C}$, respectively. By using the proposed Prototype Fusion Module, support prototype $\mathbf{P}$ is integrated with the reflectance prototype $\mathbf{P_r}$ through the co-attention mechanism. 

Updated support prototype $\mathbf{P'} \in \mathbb{R}^{1 \times 1 \times C}$ is then fed into the SSP module with updated query features $\mathbf{F'_q}$. The output is augmented prototype $\mathbf{P''} \in \mathbb{R}^{1 \times 1 \times C}$. \par
Finally, following the SSP setting, we compute the cosine distance and estimate a similarity map between the augmented prototype $\mathbf{P''}$ and query features $\mathbf{F'_q}$ to generate the final predictions $\mathbf{\hat y} \in \mathbb{R}^{H \times W \times 1}$:
\begin{equation} \label{eq:output}
\mathbf{\hat y} = \mathrm{softmax}(\mathrm{cosine}(\mathbf{P''}, \mathbf{F'_q}))
\end{equation}

\subsection{Prototype Fusion Module} \label{pfm}
A Prototype Fusion Module (PFM) is introduced to interactively fuse the support prototype $\mathbf{P}$ and reflectance prototype $\mathbf{P_{r}}$. This fusion is achieved through the co-attention mechanism. Specifically, it learns attention weights from both prototypes and integrates the weights into the prototype pair. By using our proposed Prototype Fusion Module, we effectively fuse the features extracted from different representations that carry complementary information. Such fusion results in better enhanced images in terms of improved image quality and visual perception. \par

Given the prototype $\mathbf{P}$ and the reflectance prototype $\mathbf{P_{r}}$, the concatenated features of $\mathbf{P}$ and $\mathbf{P_r}$ are fed to a convolutional layer. Next, we apply normalization on the features and use two fully connected layers to learn the attention weights as follows:
\begin{equation} \label{eq:weights}
\mathbf{W} = \mathrm{sigmoid}(f_2(\sigma(f_1(\mathbf{X}))))
\end{equation}
where $\mathbf{X} \in \mathbb{R}^{1 \times 1 \times 2C}$ represents the concatenated prototype features after normalization, $f$ represents fully connected layers and $\sigma$ represents the activation function. \par

Then, we update the support and reflectance prototypes using the following equations: 
\begin{equation} \label{eq:updatergb}
\mathbf{P'} = (1+\alpha \mathbf{W}\otimes \mathbf{P})
\end{equation}
\begin{equation} \label{eq:updateref}
\mathbf{P'_r} = (1+\alpha \mathbf{W}\otimes \mathbf{P_r})
\end{equation}
where $\alpha$ is a learnable parameter and $\otimes$ represents the Hadamard product. \par

\subsection{ASPP Module} \label{aspp}
We apply the Atrous Spatial Pyramid Pooling (ASPP) module based on the work of DeepLabV3 \cite{chen2017rethinking}. The ASPP module has atrous convolutions at multiple dilation rates and therefore, captures contextual information at various scales, addressing both local and global contexts. By incorporating multi-scale features, the ASPP module helps the network distinguish between objects of varying sizes and complex scenes with diverse textures. \par

In our work, the ASPP module is employed to capture multi-scale information from high-level query features $\mathbf{F_q}$. The resulting feature map is then upsampled and concatenated with the reflectance features $\mathbf{F_r}$ (referred as low-level features). The reflectance features help in preserving edge details, which is essential for accurately separating cracks and backgrounds. Finally, the concatenated features $\mathbf{F'_q}$ are fed to the SSP module as the second input to update the support prototype $\mathbf{F'_r}$. It is also used in generating the final matching predictions. \par

\subsection{Loss}
We train our model for the final prediction under supervision:
\begin{equation} \label{eq:mainloss}
L_{seg} = \mathrm{BCE}(\mathbf{\hat y}, \mathbf{M_q})
\end{equation}
where $\mathrm{BCE}$ is the binary cross-entropy loss, $\hat y$ is the final segmentation prediction and $\mathbf{M_q}$ is the ground truth label of the query image. $L_{seg}$ ensures that the predictions are consistent with the ground truth label. \par

To further facilitate the SSP matching procedure, we apply the self-support loss mentioned in SSP \cite{fan2022self} to measure the support and reflectance prototypes:
\begin{multline} \label{eq:supportloss}
L_{s} = \mathrm{BCE}(\mathrm{cosine}(\mathbf{P'}, \mathbf{F_{s,s}}), \mathbf{M_s}) \\ 
+\mathrm{BCE}(\mathrm{cosine}(\mathbf{P'_{r}}, \mathbf{F_{s,r}}),\mathbf{M_s})
\end{multline}

We apply the same procedure to the query features to introduce the query self-support loss $L_q$:
\begin{equation} \label{eq:queryloss}
L_{q} = \mathrm{BCE}(\mathrm{cosine}(\mathrm{MAP}(\mathbf{F'_q}), \mathbf{F'_q}), \mathbf{M_q}) 
\end{equation}
where $\mathrm{MAP}$ is the Masked Average Pooling layer used to generate a prototype on query features. \par

Finally, we train our model by optimizing all aforementioned losses jointly:
\begin{equation} \label{eq:loss}
L = L_{seg} + \lambda_1 L_s + \lambda_2 L_q
\end{equation}
where $\lambda_1$ = 1.0, $\lambda_2$ = 0,2 are the loss weights. \par

%% file: Body/4_Results.tex
\section{Experiments\label{expe}}
In this section, we describe the experiments we conduct to compare CrackNex and SOTA few-shot segmentation models using two datasets, namely (a) ll\_CrackSeg9k and (b) LCSD datasets. Our results show that our model design achieves better performance than state-of-the-art models. We also provide a detailed analysis of our design features via several ablation studies. \par

\subsection{Datasets}
\subsubsection{\textup{\textbf{ll\_CrackSeg9k}}}
The CrackSeg9k dataset \cite{kulkarni2022crackseg9k} is a popular crack-related dataset that researchers used. We select 9000 crack images from CrackSeg9k as our training set and another 1500 crack images as our test set. Since 1500 images in the test set are considered normal light images, we use Restormer \cite{zamir2022restormer} pre-trained on LDIS \cite{ying2022delving} dataset to convert them into synthetic low-light images \par

\subsubsection{\textup{\textbf{LCSD}}}
To evaluate our method under real-world lowlight conditions, we additionally collect our own crack dataset, LCSD, with 102 well-illuminated crack images as the training set and 41 low-light crack images as the test set within the Lehigh University campus. All images are taken by iPad Pro 1st generation and are resized to $400 \times 400$ for efficiency. We further annotate each crack image pixel-wise and generate a binary label. \par

\vspace{12cm}
\begin{table}[t]
\caption{Baseline comparisons on the ll\_CrackSeg9k and LCSD dataset in terms of mIOU$\uparrow$}
\normalsize
\center
\resizebox{0.5\textwidth}{!}{ 
  \begin{tabular}{c|c|cc|cc}
  \Xhline{0.8px}
  \multirow{2}{*}{Method} & \multirow{2}{*}{Backbone} & \multicolumn{2}{c|}{\textbf{CrackSeg9k}} & \multicolumn{2}{c}{\textbf{LCSD}} \\
  \cline{3-6}
  & & 1-shot  & 5-shot & 1-shot  & 5-shot \\
  \hline
   VAT \cite{hong2022cost} & \multirow{4}{*}{ResNet50} & 54.32 & 57.45 & 54.28 & 56.53 \\
   MLC \cite{yang2021mining} & & 56.54 & 58.72 & 55.48 & 57.41 \\
   SSP \cite{fan2022self} & & 60.42 & 64.25 & 56.41 & 63.30 \\    
  \textbf{CrackNex (Ours)} & & \textbf{63.00} & \textbf{69.66} & \textbf{63.85} & \textbf{65.17}\\
  \hline
   VAT \cite{hong2022cost} & \multirow{4}{*}{ResNet101} & 59.83 & 61.27 & 55.25 & 59.24 \\
   MLC \cite{yang2021mining} & & 56.73 & 62.99 & 57.18 & 58.11 \\
   SSP \cite{fan2022self} & & 56.45 & 65.29 & 56.61 & 63.16 \\    
  \textbf{CrackNex (Ours)} & & \textbf{65.90} & \textbf{70.59} & \textbf{66.10} & \textbf{68.82}\\
  \Xhline{0.8px}
\end{tabular}}
\label{result}
\vspace{-0.2cm}
\end{table}

\subsection{Implementation Details}
For the backbone CNN, We adopt the ResNet50 and ResNet101 \cite{he2016deep} pre-trained on ImageNet-1K dataset \cite{deng2009imagenet}. We train the entire framework using SGD optimizer \cite{kiefer1952stochastic} with the 0.9 momentum. The initial learning rate is 10e-3 and decayed by 10 times every 2,000 iterations. Our network is trained on one single NVIDIA TITAN RTX GPU for 6000 iterations with a batch size of 4. Both images and masks are augmented with random horizontal flipping while the evaluation is performed on the original image. \par

For comparison with our proposed scheme, we additionally evaluate the performance of several SOTA methods on the LCSD dataset. \par

\subsection{Quantitative Results}
In terms of the evaluation metrics, we use the popular mean Intersection-over-Union (mIOU$\uparrow$) to evaluate our model under 1-shot and 5-shot settings. We evaluate performance on both ll\_CrackSeg9k and LCSD benchmarks. The main results are presented in the Table. \ref{result} where we compare the performance of CrackNex with other state-of-the-art methods using the ll\_CrackSeg9k and LCSD datasets. We test SOTA methods (VAT \cite{hong2022cost}, MLC \cite{yang2021mining} and SSP \cite{fan2022self}) using their default settings. From the table, we see that CrackNex achieves an mIOU of 63.00 and 69.66 respectively under 1-shot and 5-shot settings using the ResNet50 backbone and 65.90 and 70.59 using the ResNet101 backbone \cite{he2016deep}. It outperforms SOTA methods on the ll\_CrackSeg9k dataset. \par

We additionally compare the performance of CrackNex with other state-of-the-art methods on the LCSD dataset. From the table, we found that CrackNex achieves an mIOU of 63.85 and 65.17 respectively under 1-shot and 5-shot settings using the ResNet50 backbone and 66.10 and 68.82 using the ResNet101 backbone \cite{he2016deep}. It outperforms other SOTA models by a large margin. \par

\begin{figure}
\centering
\begin{tabular}{c}
\rotatebox{90}{Query Image}
\begin{minipage}[t]{0.49\textwidth}
\includegraphics[height=1.7cm]{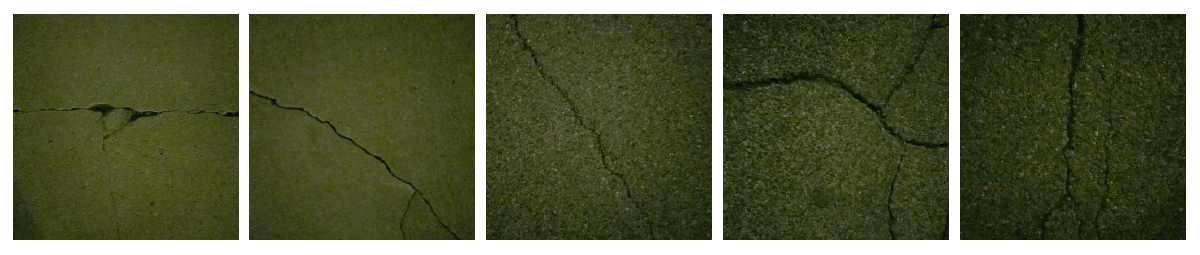}
\end{minipage} \\
\rotatebox{90}{Ground Truth}
\begin{minipage}[t]{0.49\textwidth}
\includegraphics[height=1.7cm]{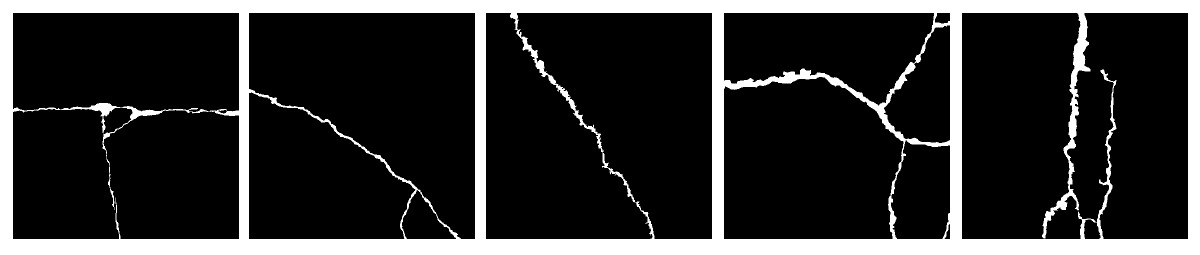}
\end{minipage} \\
\rotatebox[origin=c]{90}{Ours}
\begin{minipage}{0.49\textwidth}
\includegraphics[height=1.7cm]{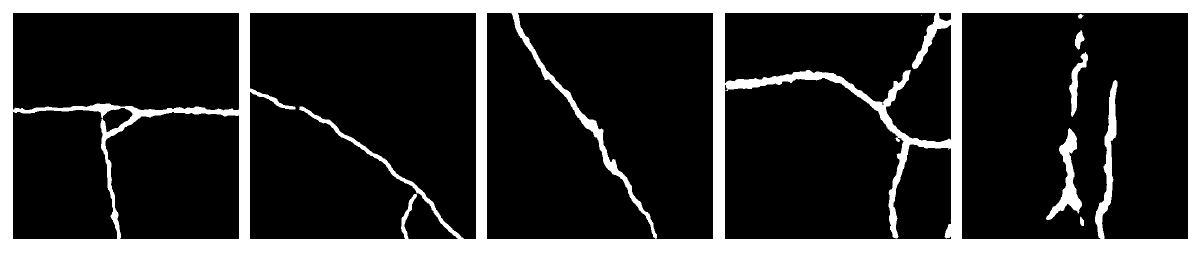}
\end{minipage} \\
\rotatebox[origin=c]{90}{MLC \cite{yang2021mining}}
\begin{minipage}{0.49\textwidth}
\includegraphics[height=1.7cm]{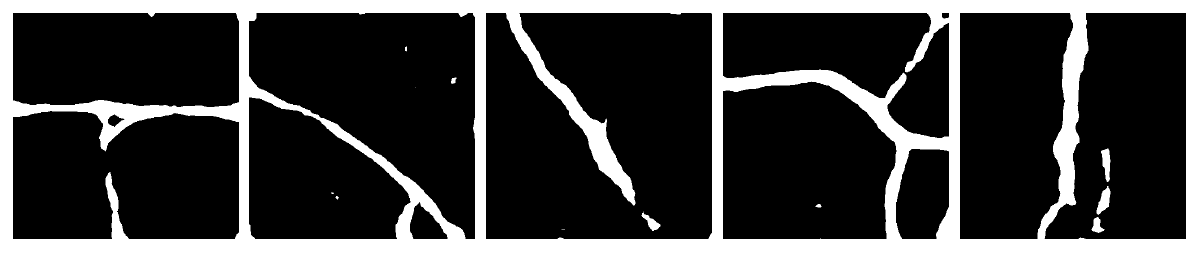}
\end{minipage} \\
\rotatebox[origin=c]{90}{SSP \cite{fan2022self}}
\begin{minipage}{0.49\textwidth}
\includegraphics[height=1.7cm]{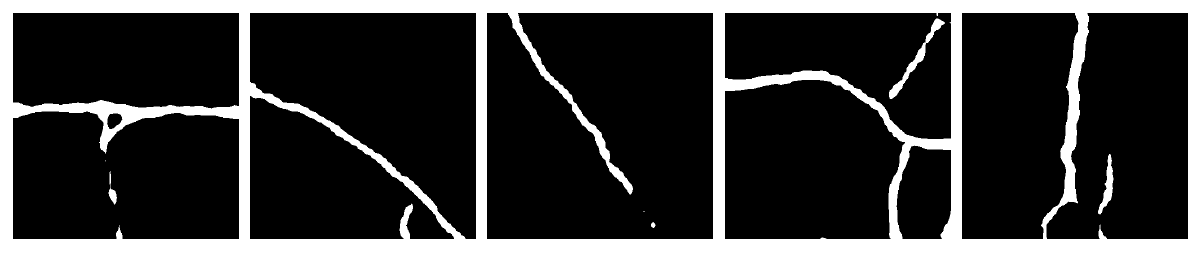}
\end{minipage}
\end{tabular} 
\caption{Qualitative results on LCSD dataset. Zoom in for details. Images have been brightened to improve their visibility.}
\label{vis}
\end{figure}

\begin{figure*}
\begin{tabular}[width=0.8\textwidth]{c}
\rotatebox[origin=c]{90}{Query Image}
\begin{minipage}{0.49\textwidth}
\includegraphics[height=3.4cm]{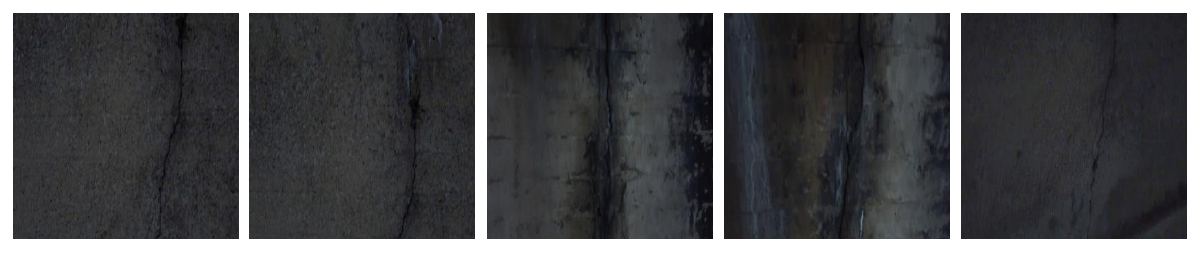}
\end{minipage} \\
\rotatebox[origin=c]{90}{Prediction}
\begin{minipage}{0.49\textwidth}
\includegraphics[height=3.4cm]{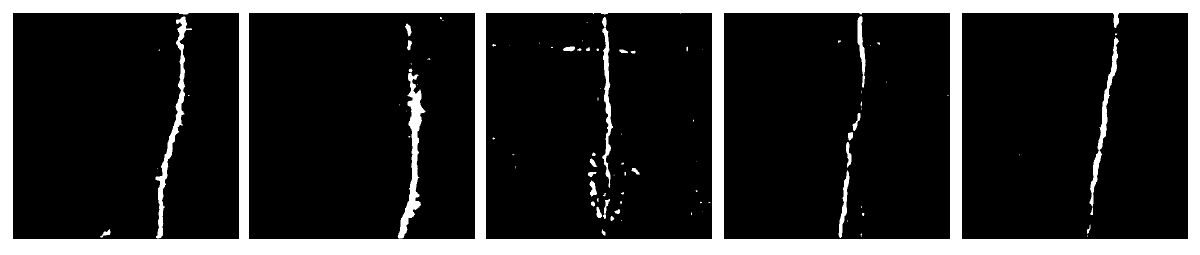}
\end{minipage}
\end{tabular} 
\caption{Qualitative results on UAV-based real-site low-light bridge cracks. Images have been brightened to improve their visibility.}
\label{uavimage}
\end{figure*}

\subsection{Qualitative Results}
To better analyze our proposed model, we visualize several segmentation results from the LCSD dataset, as shown in Fig. \ref{vis}. We compare the qualitative results of our method with two SOTA models, SSP \cite{fan2022self} and MLC \cite{yang2021mining} using their default settings. We can see that CrackNex generates more accurate boundaries and more discriminative cracks compared with existing SOTA approaches, which demonstrates the effectiveness of CrackNex. \par

\begin{table}[t]
\caption{CrackNex compared with data-driven crack segmentation models on the ll\_CrackSeg9k dataset}
\normalsize
\center
\resizebox{0.48\textwidth}{!}{ 
\fontsize{3pt}{3pt}\selectfont
\renewcommand{\arraystretch}{1.2}
  \begin{tabular}{cc}
  \Xhline{0.25px}
  Method & mIOU$\uparrow$ \\
  \Xhline{0.1px}
   DDRNet \cite{peng2022pp} & 69.23 \\
   DeepLabV3 \cite{chen2017rethinking} & 71.16 \\
   STDCSeg \cite{fan2021rethinking} & 70.98 \\
   HrSegNet-B16 \cite{li2023real} & 71.32 \\ 
   HrSegNet-B32 \cite{li2023real} & 72.45 \\    
   \Xhline{0.1px}
  \textbf{CrackNex (Ours)} & 70.59 \\
  \Xhline{0.25px}
\end{tabular}}
\label{compare}
\vspace{-0.2cm}
\end{table}

To further evaluate the UAV inspection system in real-world scenarios, we use a drone to conduct the inspections of a nearby bridge built in 1924. As shown in Fig. \ref{uavimage}, the segmentation results are accurate and demonstrate the effectiveness of our method. \par

\subsection{Compared with data-driven crack segmentation models}
We additionally compare our model with several SOTA data-driven crack semantic segmentation models on the ll\_CrackSeg9k dataset. The main results are presented in the Table. \ref{compare}. We test SOTA methods (DDRNet \cite{peng2022pp}, DeepLabV3 \cite{chen2017rethinking}, STDCSeg \cite{fan2021rethinking} and HrSegNet \cite{li2023real}) using their default settings. From the table, we see that CrackNex achieves mIOU of 70.59 under 5-shot settings and achieves comparable results compared with these data-driven crack segmentation models. Note that these data-driven methods require a large amount of labeled low-light crack images to achieve optimal performance, whereas our model doesn't need to be trained on low-light crack images. \par

\subsection{Ablation Study}
We further perform ablation studies on the LCSD dataset using the ResNet101 backbone \cite{he2016deep} to investigate the contribution of key technical components in our method. \par 

Specifically, we evaluate four variants of CrackNex: (i) baseline architecture, (ii) adding reflectance features, (iii) utilization of Prototype Fusion Module (PFM), and (iv) employment of all the designed components. The results of the ablation study are summarized in Table \ref{ablation}.

\begin{table*}[t]
\centering
\resizebox{0.5\textwidth}{!}{
  \begin{tabular}{ccc|cc}
  \Xhline{0.8px}
  Reflectance & PFM & ASPP & \multicolumn{2}{c}{mIOU$\uparrow$} \\
  features & Module & module & 1-shot & 5-shot \\
  \hline
   &  &  & 56.61 & 63.16 \\
  \checkmark &  &  & 63.33 & 65.79 \\
  \checkmark & \checkmark &  & 65.06 & 67.37 \\
  \checkmark & \checkmark & \checkmark & \textbf{66.10} & \textbf{68.82} \\
  \Xhline{0.8px}
    \end{tabular}
}
\caption{Ablation study of adding different components on LCSD dataset}
\label{ablation}
\end{table*}

The second row of Table \ref{ablation} investigates the effectiveness of reflectance features in CrackNex. We directly concatenate support features and support reflectance features and feed the concatenated features to generate one single support prototype. Our results demonstrate that incorporating the reflectance features can significantly enhance performance. \par

We further test a variant “w/ PFM” where we utilize the PFM module to generate dual prototypes. Instead of early concatenating support features and support reflectance features, we use a co-attention mechanism to interactively update the support prototype. The results are reported in row 3. Experimental results show that PFM helps in generating better prototypes. \par

As for the ASPP module, our experimental results (compared row 3 with 4) demonstrate that adding the ASPP module provides further feature extraction capabilities and yields better performance. \par

\section{Conclusion\label{conclusion}}
In this paper, we propose a novel reflectance-guided few-shot low-light crack segmentation framework, CrackNex. We utilize few-shot segmentation to solve the problem of having to annotate many training images. In addition, we introduce reflectance information to improve segmentation predictions during low-light environments. We validate our framework on two low-light crack datasets, ll\_CrackSeg9k and LCSD, and demonstrate significant improvements in the mIOU metric. Our results highlight the importance of incorporating reflectance features to capture details and enhance object boundaries. Additionally, we release a new crack dataset with both well-illuminated and low-light crack images for the benefit of the research community. \par

\section*{Acknowledgments}
This work was supported by National Science Foundation Grant CPS 1931867. We would also like to express our gratitude to the anonymous reviewers for their insightful comments and suggestions. \par

%% file: main.bbl
\begin{thebibliography}{10}
\providecommand{\url}[1]{#1}
\csname url@samestyle\endcsname
\providecommand{\newblock}{\relax}
\providecommand{\bibinfo}[2]{#2}
\providecommand{\BIBentrySTDinterwordspacing}{\spaceskip=0pt\relax}
\providecommand{\BIBentryALTinterwordstretchfactor}{4}
\providecommand{\BIBentryALTinterwordspacing}{\spaceskip=\fontdimen2\font plus
\BIBentryALTinterwordstretchfactor\fontdimen3\font minus
  \fontdimen4\font\relax}
\providecommand{\BIBforeignlanguage}[2]{{%
\expandafter\ifx\csname l@#1\endcsname\relax
\typeout{** WARNING: IEEEtran.bst: No hyphenation pattern has been}%
\typeout{** loaded for the language `#1'. Using the pattern for}%
\typeout{** the default language instead.}%
\else
\language=\csname l@#1\endcsname
\fi
#2}}
\providecommand{\BIBdecl}{\relax}
\BIBdecl

\bibitem{yan2022cycleadc}
Y.~Yan, S.~Zhu, S.~Ma, Y.~Guo, and Z.~Yu, ``Cycleadc-net: A crack segmentation
  method based on multi-scale feature fusion,'' \emph{Measurement}, vol. 204,
  p. 112107, 2022.

\bibitem{choi2019sddnet}
W.~Choi and Y.-J. Cha, ``Sddnet: Real-time crack segmentation,'' \emph{IEEE
  Transactions on Industrial Electronics}, vol.~67, no.~9, pp. 8016--8025,
  2019.

\bibitem{liu2019deepcrack}
Y.~Liu, J.~Yao, X.~Lu, R.~Xie, and L.~Li, ``Deepcrack: A deep hierarchical
  feature learning architecture for crack segmentation,''
  \emph{Neurocomputing}, vol. 338, pp. 139--153, 2019.

\bibitem{kang2020hybrid}
D.~Kang, S.~S. Benipal, D.~L. Gopal, and Y.-J. Cha, ``Hybrid pixel-level
  concrete crack segmentation and quantification across complex backgrounds
  using deep learning,'' \emph{Automation in Construction}, vol. 118, p.
  103291, 2020.

\bibitem{liu2020automated}
J.~Liu, X.~Yang, S.~Lau, X.~Wang, S.~Luo, V.~C.-S. Lee, and L.~Ding,
  ``Automated pavement crack detection and segmentation based on two-step
  convolutional neural network,'' \emph{Computer-Aided Civil and Infrastructure
  Engineering}, vol.~35, no.~11, pp. 1291--1305, 2020.

\bibitem{rezaie2020comparison}
A.~Rezaie, R.~Achanta, M.~Godio, and K.~Beyer, ``Comparison of crack
  segmentation using digital image correlation measurements and deep
  learning,'' \emph{Construction and Building Materials}, vol. 261, p. 120474,
  2020.

\bibitem{zheng2023robustness}
Z.~Zheng, X.~Ying, Z.~Yao, and M.~C. Chuah, ``Robustness of trajectory
  prediction models under map-based attacks,'' in \emph{Proceedings of the
  IEEE/CVF Winter Conference on Applications of Computer Vision}, 2023, pp.
  4541--4550.

\bibitem{hsieh2021neural}
Y.-T. Hsieh, K.~Anjum, S.~Huang, I.~Kulkarni, and D.~Pompili, ``Neural network
  design via voltage-based resistive processing unit and diode activation
  function-a new architecture,'' in \emph{2021 IEEE International Midwest
  Symposium on Circuits and Systems (MWSCAS)}.\hskip 1em plus 0.5em minus
  0.4em\relax IEEE, 2021, pp. 59--62.

\bibitem{wang2024landa}
Z.~Wang, L.~Zhang, L.~Wang, and M.~Zhu, ``Landa: Language-guided multi-source
  domain adaptation,'' \emph{arXiv preprint arXiv:2401.14148}, 2024.

\bibitem{zhu2023topic}
Y.~Zhu, Y.~Qiu, Q.~Wu, F.~L. Wang, and Y.~Rao, ``Topic driven adaptive network
  for cross-domain sentiment classification,'' \emph{Information Processing \&
  Management}, vol.~60, no.~2, p. 103230, 2023.

\bibitem{wang2022automatic}
W.~Wang and C.~Su, ``Automatic concrete crack segmentation model based on
  transformer,'' \emph{Automation in Construction}, vol. 139, p. 104275, 2022.

\bibitem{fan2023pavement}
L.~Fan, S.~Li, Y.~Li, B.~Li, D.~Cao, and F.-Y. Wang, ``Pavement cracks coupled
  with shadows: A new shadow-crack dataset and a shadow-removal-oriented crack
  detection approach,'' \emph{IEEE/CAA Journal of Automatica Sinica}, vol.~10,
  no.~7, pp. 1593--1607, 2023.

\bibitem{li2019crack}
W.~Li, Z.~Shen, and P.~Li, ``Crack detection of track plate based on yolo,'' in
  \emph{2019 12th international symposium on computational intelligence and
  design (ISCID)}, vol.~2.\hskip 1em plus 0.5em minus 0.4em\relax IEEE, 2019,
  pp. 15--18.

\bibitem{li2021automatic}
G.~Li, Q.~Liu, W.~Ren, W.~Qiao, B.~Ma, and J.~Wan, ``Automatic recognition and
  analysis system of asphalt pavement cracks using interleaved low-rank group
  convolution hybrid deep network and segnet fusing dense condition random
  field,'' \emph{Measurement}, vol. 170, p. 108693, 2021.

\bibitem{wang2021renet}
Y.~Wang, K.~Song, J.~Liu, H.~Dong, Y.~Yan, and P.~Jiang, ``Renet: Rectangular
  convolution pyramid and edge enhancement network for salient object detection
  of pavement cracks,'' \emph{Measurement}, vol. 170, p. 108698, 2021.

\bibitem{dong2022innovative}
J.~Dong, N.~Wang, H.~Fang, Q.~Hu, C.~Zhang, B.~Ma, D.~Ma, and H.~Hu,
  ``Innovative method for pavement multiple damages segmentation and
  measurement by the road-seg-capsnet of feature fusion,'' \emph{Construction
  and Building Materials}, vol. 324, p. 126719, 2022.

\bibitem{land1977retinex}
E.~H. Land, ``The retinex theory of color vision,'' \emph{Scientific american},
  vol. 237, no.~6, pp. 108--129, 1977.

\bibitem{wei2018deep}
C.~Wei, W.~Wang, W.~Yang, and J.~Liu, ``Deep retinex decomposition for
  low-light enhancement,'' \emph{arXiv preprint arXiv:1808.04560}, 2018.

\bibitem{zhao2021few}
N.~Zhao, T.-S. Chua, and G.~H. Lee, ``Few-shot 3d point cloud semantic
  segmentation,'' in \emph{Proceedings of the IEEE/CVF Conference on Computer
  Vision and Pattern Recognition}, 2021, pp. 8873--8882.

\bibitem{wang2019panet}
K.~Wang, J.~H. Liew, Y.~Zou, D.~Zhou, and J.~Feng, ``Panet: Few-shot image
  semantic segmentation with prototype alignment,'' in \emph{proceedings of the
  IEEE/CVF international conference on computer vision}, 2019, pp. 9197--9206.

\bibitem{dong2018few}
N.~Dong and E.~P. Xing, ``Few-shot semantic segmentation with prototype
  learning.'' in \emph{BMVC}, vol.~3, no.~4, 2018.

\bibitem{min2021hypercorrelation}
J.~Min, D.~Kang, and M.~Cho, ``Hypercorrelation squeeze for few-shot
  segmentation,'' in \emph{Proceedings of the IEEE/CVF international conference
  on computer vision}, 2021, pp. 6941--6952.

\bibitem{mao2022learning}
B.~Mao, X.~Zhang, L.~Wang, Q.~Zhang, S.~Xiang, and C.~Pan, ``Learning from the
  target: Dual prototype network for few shot semantic segmentation,'' in
  \emph{Proceedings of the AAAI Conference on Artificial Intelligence},
  vol.~36, no.~2, 2022, pp. 1953--1961.

\bibitem{sun2022few}
L.~Sun, C.~Li, X.~Ding, Y.~Huang, Z.~Chen, G.~Wang, Y.~Yu, and J.~Paisley,
  ``Few-shot medical image segmentation using a global correlation network with
  discriminative embedding,'' \emph{Computers in biology and medicine}, vol.
  140, p. 105067, 2022.

\bibitem{pan2023learning}
P.~Pan, Z.~Fan, B.~Y. Feng, P.~Wang, C.~Li, and Z.~Wang, ``Learning to estimate
  6dof pose from limited data: A few-shot, generalizable approach using rgb
  images,'' \emph{arXiv preprint arXiv:2306.07598}, 2023.

\bibitem{chen2017rethinking}
L.-C. Chen, G.~Papandreou, F.~Schroff, and H.~Adam, ``Rethinking atrous
  convolution for semantic image segmentation,'' \emph{arXiv preprint
  arXiv:1706.05587}, 2017.

\bibitem{finn2017model}
C.~Finn, P.~Abbeel, and S.~Levine, ``Model-agnostic meta-learning for fast
  adaptation of deep networks,'' in \emph{International conference on machine
  learning}.\hskip 1em plus 0.5em minus 0.4em\relax PMLR, 2017, pp. 1126--1135.

\bibitem{garcia2017few}
V.~Garcia and J.~Bruna, ``Few-shot learning with graph neural networks,''
  \emph{arXiv preprint arXiv:1711.04043}, 2017.

\bibitem{munkhdalai2017meta}
T.~Munkhdalai and H.~Yu, ``Meta networks,'' in \emph{International conference
  on machine learning}.\hskip 1em plus 0.5em minus 0.4em\relax PMLR, 2017, pp.
  2554--2563.

\bibitem{wang2022metateacher}
Z.~Wang, M.~Ye, X.~Zhu, L.~Peng, L.~Tian, and Y.~Zhu, ``Metateacher:
  Coordinating multi-model domain adaptation for medical image
  classification,'' \emph{Advances in Neural Information Processing Systems},
  vol.~35, pp. 20\,823--20\,837, 2022.

\bibitem{li2022domain}
C.~Li, X.~Lin, Y.~Mao, W.~Lin, Q.~Qi, X.~Ding, Y.~Huang, D.~Liang, and Y.~Yu,
  ``Domain generalization on medical imaging classification using episodic
  training with task augmentation,'' \emph{Computers in biology and medicine},
  vol. 141, p. 105144, 2022.

\bibitem{elsken2020meta}
T.~Elsken, B.~Staffler, J.~H. Metzen, and F.~Hutter, ``Meta-learning of neural
  architectures for few-shot learning,'' in \emph{Proceedings of the IEEE/CVF
  conference on computer vision and pattern recognition}, 2020, pp.
  12\,365--12\,375.

\bibitem{zoph2016neural}
B.~Zoph and Q.~V. Le, ``Neural architecture search with reinforcement
  learning,'' \emph{arXiv preprint arXiv:1611.01578}, 2016.

\bibitem{baik2021meta}
S.~Baik, J.~Choi, H.~Kim, D.~Cho, J.~Min, and K.~M. Lee, ``Meta-learning with
  task-adaptive loss function for few-shot learning,'' in \emph{Proceedings of
  the IEEE/CVF international conference on computer vision}, 2021, pp.
  9465--9474.

\bibitem{ying2021weakly}
X.~Ying, X.~Li, and M.~C. Chuah, ``Weakly-supervised object representation
  learning for few-shot semantic segmentation,'' in \emph{Proceedings of the
  IEEE/CVF Winter Conference on Applications of Computer Vision}, 2021, pp.
  1497--1506.

\bibitem{fan2022self}
Q.~Fan, W.~Pei, Y.-W. Tai, and C.-K. Tang, ``Self-support few-shot semantic
  segmentation,'' in \emph{European Conference on Computer Vision}.\hskip 1em
  plus 0.5em minus 0.4em\relax Springer, 2022, pp. 701--719.

\bibitem{okazawa2022interclass}
A.~Okazawa, ``Interclass prototype relation for few-shot segmentation,'' in
  \emph{European Conference on Computer Vision}.\hskip 1em plus 0.5em minus
  0.4em\relax Springer, 2022, pp. 362--378.

\bibitem{lau2020automated}
S.~L. Lau, E.~K. Chong, X.~Yang, and X.~Wang, ``Automated pavement crack
  segmentation using u-net-based convolutional neural network,'' \emph{Ieee
  Access}, vol.~8, pp. 114\,892--114\,899, 2020.

\bibitem{konig2021optimized}
J.~K{\"o}nig, M.~D. Jenkins, M.~Mannion, P.~Barrie, and G.~Morison, ``Optimized
  deep encoder-decoder methods for crack segmentation,'' \emph{Digital Signal
  Processing}, vol. 108, p. 102907, 2021.

\bibitem{qiu2023sats}
Y.~Qiu, Y.~Shen, Z.~Sun, Y.~Zheng, X.~Chang, W.~Zheng, and R.~Wang, ``Sats:
  Self-attention transfer for continual semantic segmentation,'' \emph{Pattern
  Recognition}, vol. 138, p. 109383, 2023.

\bibitem{liu2019deep}
K.~Liu, X.~Han, and B.~M. Chen, ``Deep learning based automatic crack detection
  and segmentation for unmanned aerial vehicle inspections,'' in \emph{2019
  IEEE international conference on robotics and biomimetics (ROBIO)}.\hskip 1em
  plus 0.5em minus 0.4em\relax IEEE, 2019, pp. 381--387.

\bibitem{sun2023hmaac}
C.~Sun, S.~Huang, and D.~Pompili, ``Hmaac: Hierarchical multi-agent
  actor-critic for aerial search with explicit coordination modeling,'' in
  \emph{2023 IEEE International Conference on Robotics and Automation
  (ICRA)}.\hskip 1em plus 0.5em minus 0.4em\relax IEEE, 2023, pp. 7728--7734.

\bibitem{wei2022lidar}
Y.~Wei, Z.~Wei, Y.~Rao, J.~Li, J.~Zhou, and J.~Lu, ``Lidar distillation:
  Bridging the beam-induced domain gap for 3d object detection,'' in
  \emph{European Conference on Computer Vision}.\hskip 1em plus 0.5em minus
  0.4em\relax Springer, 2022, pp. 179--195.

\bibitem{dang2024enhancing}
B.~Dang, D.~Ma, S.~Li, X.~Dong, H.~Zang, and R.~Ding, ``Enhancing kitchen
  independence: Deep learning-based object detection for visually impaired
  assistance,'' \emph{Academic Journal of Science and Technology}, vol.~9,
  no.~2, pp. 180--184, 2024.

\bibitem{ma2023implementation}
D.~Ma, B.~Dang, S.~Li, H.~Zang, and X.~Dong, ``Implementation of computer
  vision technology based on artificial intelligence for medical image
  analysis,'' \emph{International Journal of Computer Science and Information
  Technology}, vol.~1, no.~1, pp. 69--76, 2023.

\bibitem{sarmiento2021pavement}
J.-A. Sarmiento, ``Pavement distress detection and segmentation using yolov4
  and deeplabv3 on pavements in the philippines,'' \emph{arXiv preprint
  arXiv:2103.06467}, 2021.

\bibitem{xiang2023crack}
C.~Xiang, J.~Guo, R.~Cao, and L.~Deng, ``A crack-segmentation algorithm fusing
  transformers and convolutional neural networks for complex detection
  scenarios,'' \emph{Automation in Construction}, vol. 152, p. 104894, 2023.

\bibitem{ying2020}
X.~Ying, X.~Li, and M.~C. Chuah, ``Weakly-supervised object representation
  learning for few-shot semantic segmentation,'' in \emph{Proceedings of the
  IEEE WACV}, 2020.

\bibitem{deng2009imagenet}
J.~Deng, W.~Dong, R.~Socher, L.-J. Li, K.~Li, and L.~Fei-Fei, ``Imagenet: A
  large-scale hierarchical image database,'' in \emph{2009 IEEE conference on
  computer vision and pattern recognition}.\hskip 1em plus 0.5em minus
  0.4em\relax Ieee, 2009, pp. 248--255.

\bibitem{kulkarni2022crackseg9k}
S.~Kulkarni, S.~Singh, D.~Balakrishnan, S.~Sharma, S.~Devunuri, and S.~C.~R.
  Korlapati, ``Crackseg9k: a collection and benchmark for crack segmentation
  datasets and frameworks,'' in \emph{European Conference on Computer
  Vision}.\hskip 1em plus 0.5em minus 0.4em\relax Springer, 2022, pp. 179--195.

\bibitem{zamir2022restormer}
S.~W. Zamir, A.~Arora, S.~Khan, M.~Hayat, F.~S. Khan, and M.-H. Yang,
  ``Restormer: Efficient transformer for high-resolution image restoration,''
  in \emph{Proceedings of the IEEE/CVF conference on computer vision and
  pattern recognition}, 2022, pp. 5728--5739.

\bibitem{ying2022delving}
X.~Ying, B.~Lang, Z.~Zheng, and M.~C. Chuah, ``Delving into light-dark semantic
  segmentation for indoor scenes understanding,'' in \emph{Proceedings of the
  1st Workshop on Photorealistic Image and Environment Synthesis for Multimedia
  Experiments}, 2022, pp. 3--9.

\bibitem{hong2022cost}
S.~Hong, S.~Cho, J.~Nam, S.~Lin, and S.~Kim, ``Cost aggregation with 4d
  convolutional swin transformer for few-shot segmentation,'' in \emph{European
  Conference on Computer Vision}.\hskip 1em plus 0.5em minus 0.4em\relax
  Springer, 2022, pp. 108--126.

\bibitem{yang2021mining}
L.~Yang, W.~Zhuo, L.~Qi, Y.~Shi, and Y.~Gao, ``Mining latent classes for
  few-shot segmentation,'' in \emph{Proceedings of the IEEE/CVF international
  conference on computer vision}, 2021, pp. 8721--8730.

\bibitem{he2016deep}
K.~He, X.~Zhang, S.~Ren, and J.~Sun, ``Deep residual learning for image
  recognition,'' in \emph{Proceedings of the IEEE conference on computer vision
  and pattern recognition}, 2016, pp. 770--778.

\bibitem{kiefer1952stochastic}
J.~Kiefer and J.~Wolfowitz, ``Stochastic estimation of the maximum of a
  regression function,'' \emph{The Annals of Mathematical Statistics}, pp.
  462--466, 1952.

\bibitem{peng2022pp}
J.~Peng, Y.~Liu, S.~Tang, Y.~Hao, L.~Chu, G.~Chen, Z.~Wu, Z.~Chen, Z.~Yu, Y.~Du
  \emph{et~al.}, ``Pp-liteseg: A superior real-time semantic segmentation
  model,'' \emph{arXiv preprint arXiv:2204.02681}, 2022.

\bibitem{fan2021rethinking}
M.~Fan, S.~Lai, J.~Huang, X.~Wei, Z.~Chai, J.~Luo, and X.~Wei, ``Rethinking
  bisenet for real-time semantic segmentation,'' in \emph{Proceedings of the
  IEEE/CVF conference on computer vision and pattern recognition}, 2021, pp.
  9716--9725.

\bibitem{li2023real}
Y.~Li, R.~Ma, H.~Liu, and G.~Cheng, ``Real-time high-resolution neural network
  with semantic guidance for crack segmentation,'' \emph{Automation in
  Construction}, vol. 156, p. 105112, 2023.

\end{thebibliography}
